%% file: main.tex
\documentclass{article}
\usepackage{spconf,amsmath,epsfig}
\usepackage{amssymb}

\usepackage{color,soul}
\usepackage{algorithmic}
\usepackage{graphicx}
\usepackage{textcomp}
\usepackage{adjustbox}
\usepackage{subfig}
\usepackage{booktabs}
\usepackage{rotating}
\usepackage{hyperref}
\usepackage[table,x11names]{xcolor}
\hypersetup{colorlinks,allcolors=black}

\title{Land Cover Segmentation with Sparse Annotations from Sentinel-2 Imagery}
\name{Marco Galatola$^{1}$, Edoardo Arnaudo$^{1,2}$, Luca Barco$^1$, Claudio Rossi$^1$, Fabrizio Dominici$^1$
    \thanks{This publication is part of the project NODES which has received funding from the MUR – M4C2 1.5 of PNRR with grant agreement no. ECS00000036. This work is also part of the H2020 projects: SAFERS (GA n.869353) and OVERWATCH (GA n.101082320).}
}
\address{
    1. LINKS Foundation, \textit{AI, Data \& Space (ADS)}, Torino (TO), Italy \\
    2. Politecnico di Torino, \textit{Dipartimento di Automatica e Informatica  (DAUIN)}, Torino (TO), Italy
}
\begin{document}
%
\maketitle
\begin{abstract}
\input{sections/00-abstract}
\end{abstract}
\section{Introduction}
\label{sec:intro}
\input{sections/01-intro}

\section{Related Work}
\label{sec:related}
\input{sections/02-related}

\section{Dataset}
\label{sec:dataset}
\input{sections/03-dataset}

\section{Method}
\label{sec:method}
\input{sections/04-method}

\section{Experiments}
\label{sec:experiments}
\input{sections/05-experiments}

\section{Conclusions}
\label{sec:conclusion}
\input{sections/06-conclusions}

\bibliographystyle{IEEEbib}
{\footnotesize
\bibliography{IEEEabrv,refs.bib}
}

\end{document}

%% file: sections/00-abstract.tex
Land cover (LC) segmentation plays a critical role in various applications, including environmental analysis and natural disaster management. However, generating accurate LC maps is a complex and time-consuming task that requires the expertise of multiple annotators and regular updates to account for environmental changes. In this work, we introduce SPADA, a framework for fuel map delineation that addresses the challenges associated with LC segmentation using sparse annotations and domain adaptation techniques for semantic segmentation. Performance evaluations using reliable ground truths, such as LUCAS and Urban Atlas, demonstrate the technique's effectiveness. SPADA outperforms state-of-the-art semantic segmentation approaches as well as third-party products, achieving a mean Intersection over Union (IoU) score of 42.86 and an F1 score of 67.93 on Urban Atlas and LUCAS, respectively.

%% file: sections/01-intro.tex
Land Cover (LC) segmentation plays a crucial role in various applications, including urban analysis and natural disaster management \cite{hua2021semantic}.
However, the manual production of accurate maps is a time-consuming activity that often requires several expert annotators. Additionally, regular updates are necessary to account for environmental changes. In natural disaster management, such information is crucial for studying the propagation and impact of disasters such as wildfires and floods, requiring the differentiation of flammable areas (forests, shrubs) from urban borders (buildings, roads).
 
However, generating efficient and reliable LC maps introduces several unique challenges that needs to be addressed to achieve accurate results.
Existing EU-wide open LC datasets typically exhibit lower spatial resolution than required by some applications, while high-resolution products are often sparse and limited in terms of classification taxonomy.

To address these challenges, we selectively combine existing Copernicus\footnote{www.copernicus.eu} datasets, namely Corine Land Cover (CLC), Urban Atlas (UA), and Land Use and Coverage Area frame Survey (LUCAS), taking advantage of the strength of each data source.

To cope with the sparse ground truth, we propose a novel framework called \textit{SParse Annotations with DAformer} (SPADA). Leveraging Unsupervised Domain Adaptation (UDA) techniques, we propose a \textit{teacher-student} framework, where the teacher model generates robust pseudo-labels to expand the annotations across the full input space. Similar to DAFormer\cite{hoyer2022daformer}, we mix the pseudo-labels, filtered and weighted by their prediction confidence, with the processed sparse ground truth to overcome feedback loops during self-training.
We compare our solution with standard semantic segmentation approaches and third-party products, including the Sentinel-2
Global Land Cover (S2GLC), achieving a mean IoU score of 42.86 and F1 Score of 67.93 on UA and LUCAS, respectively. SPADA, without multi-temporal source images or post-processing, outperforms the strongest baselines by $+7.88$ mean IoU and $+3.86$ F1 on LUCAS.

In short, our contributions can be summarized as follows: first, we propose and evaluate the capability of SPADA, a novel framework for creating fuel maps from Sentinel-2 using vision transformers and exploiting a UDA technique leveraging on labelled and unlabeled pixels during training. Second, we release the dataset and the code used in this work, comprising the input data and sparse annotations used to train the segmentation model. 
\footnote{Code and dataset: \href{https://github.com/links-ads/spada}{https://github.com/links-ads/spada}}


%% file: sections/02-related.tex
\vspace{-1em}
In aerial and remote sensing, semantic segmentation is applied to various target environments such as urban areas \cite{hua2021semantic}, land cover \cite{s2glc2020malinowski}, and agricultural scenarios \cite{tavera2022aug_invar}.
Supervised semantic segmentation from remotely sensed imagery presents several challenges, including the high number of input bands, the image size, the top-down viewpoint, and limited ground truth availability.
To increase the segmentation performances, extra bands are typically included by introducing multiple encoders, or by expanding the input layers \cite{tavera2022aug_invar}, while the large input dimensions and the top-down viewpoint can be exploited to implement additional regularization, considering multiscale regularization \cite{chen2019glnet} or invariance to rotation \cite{tavera2022aug_invar}, both at training or test time.
In critical tasks, the presence of sparsely annotated dataset poses additional challenges to achieving acceptable performances and can be addressed with weakly-supervised approaches, which rely on less precise annotations.
In this context, Class Activation Maps (CAM) or attention maps have been effectively used \cite{ahn2018affinitynet1}, propagating labels from discriminative image regions.
Also, approaches like AffinityNet \cite{ahn2018affinitynet1} and its variants \cite{nivaggioli2019affinitynet2} predict semantic affinities between adjacent image portions, penalizing regions with different semantics. 
%
Sparse annotations are often addressed using scribbles, which provide efficient but approximate labels \cite{lin2016scribblesup}. Similar to weak supervision, the goal is to expand the sparse ground truth to every pixel of the object, ensuring semantic consistency.
Approaches like ScribbleSup \cite{lin2016scribblesup} employ graph propagation. 
Tree Energy Loss \cite{liang2022tree_energy_loss} utilizes a minimum spanning tree among pixels for pairwise affinities, while FESTA \cite{hua2021semantic} leverages an unsupervised neighbourhood loss.
In this work, we adapt the self-training UDA methods 
\cite{hoyer2022daformer}, substituting the concept of source inputs with pairs of images and sparse labels, and target inputs with the same images with scribbles mixed with pseudo-labels, as detailed in Sec. \ref{sec:method}.

%% file: sections/03-dataset.tex
\begin{figure*}
  \begin{minipage}[b]{.5\linewidth}
    \centering
        \centering
    \subfloat[Sentinel-2]{
        \label{subfig:france_s2}
        \centering
        \includegraphics[width=0.31\textwidth]{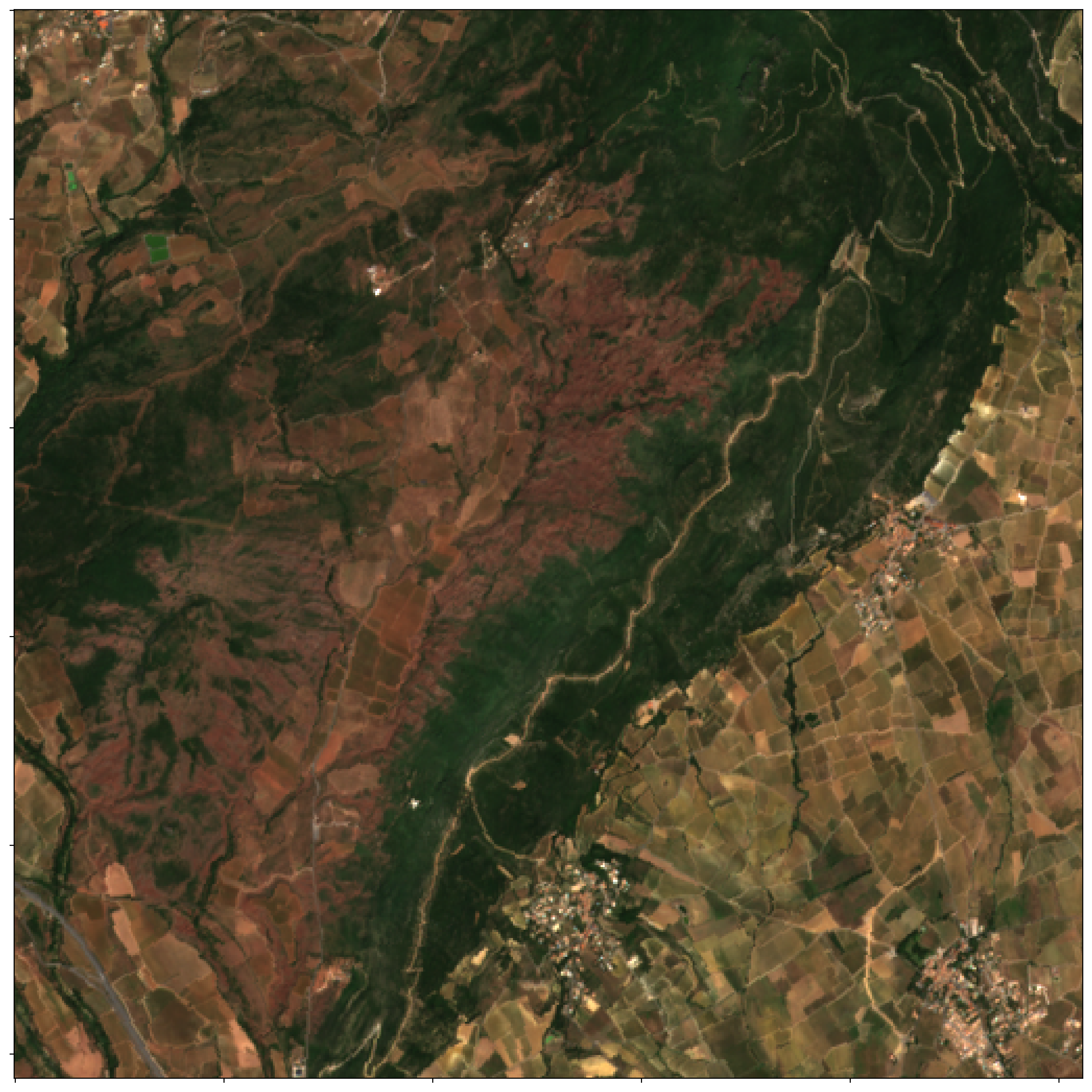}
    }
    \hfill
    \subfloat[LUCAS]{
        \label{subfig:france_lucas}
        \centering
        \includegraphics[width=0.31\textwidth]{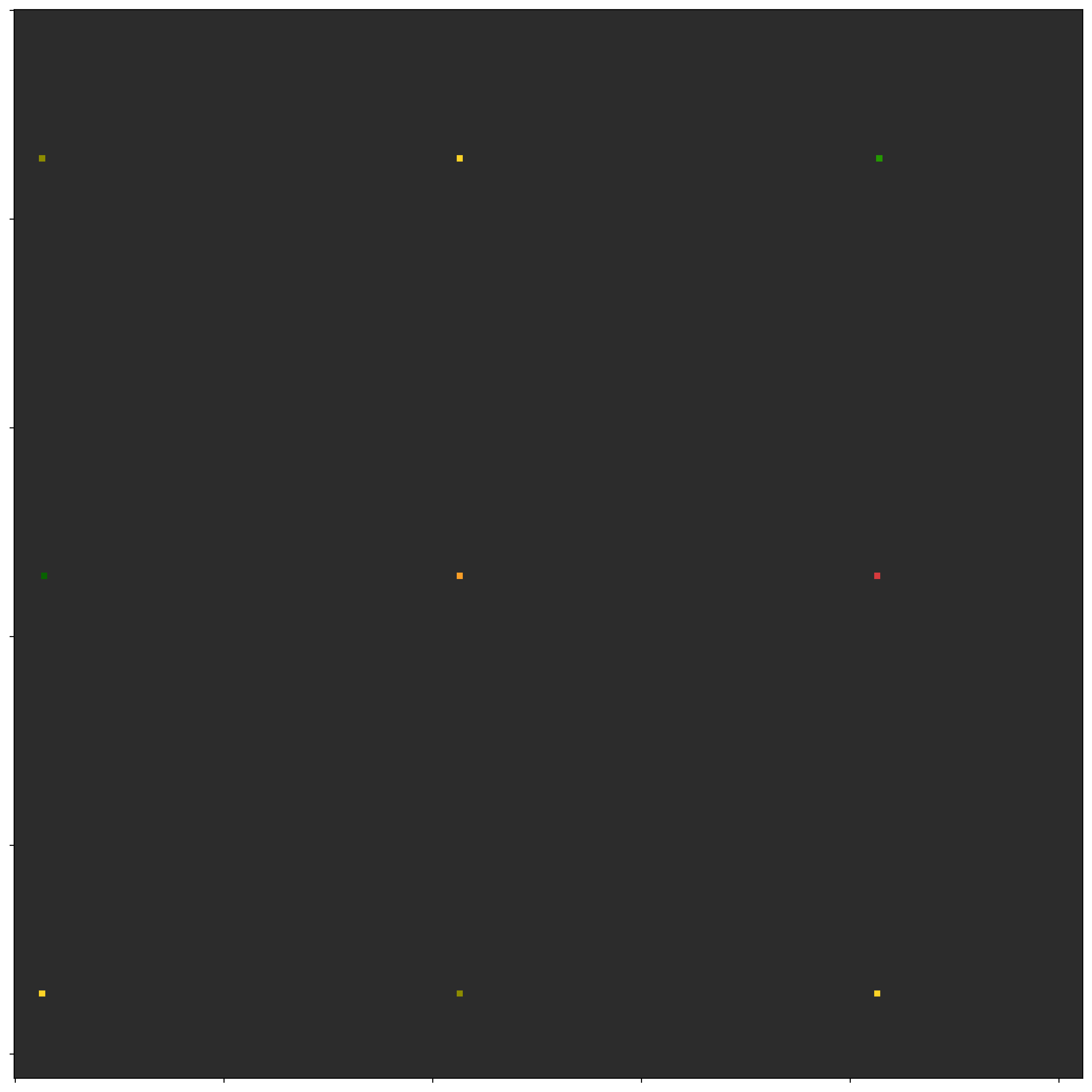}
    }
    \hfill
    \subfloat[Fuel Map scribbles]{
        \label{subfig:france_scribble_legend}
        \centering
        \includegraphics[width=0.31\textwidth]{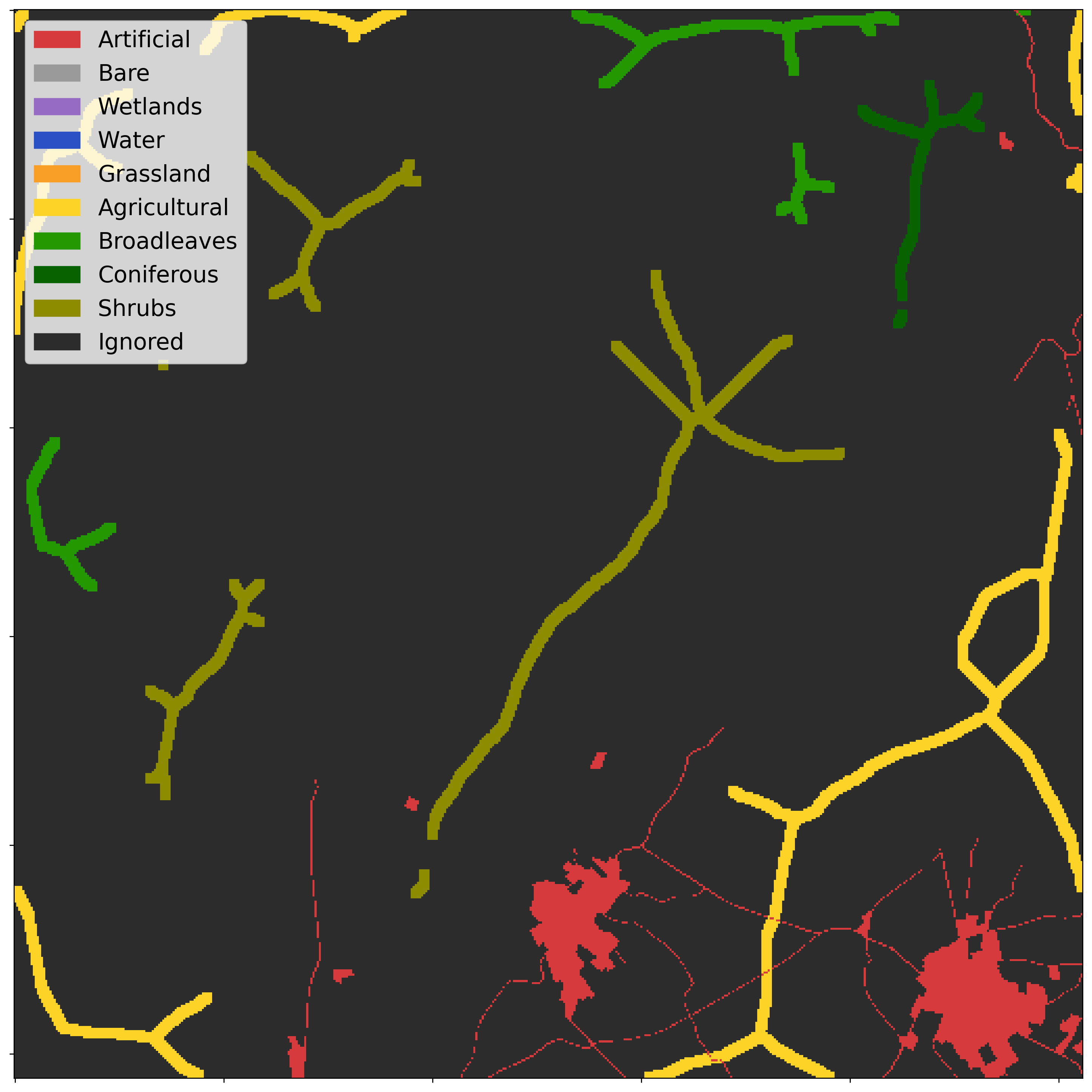}
    }
    \label{fig:dataset}
    \captionof{figure}{Input image and corresponding annotations extracted from the final fuel map dataset.}
  \end{minipage}\hfill
  \begin{minipage}[b]{.48\linewidth}
    \centering
    \input{tables/mapping}
    \captionof{table}{Mapping and aggregation carried out for CLC and and LUCAS class IDs.}
    \label{table:mapping}
  \end{minipage}
\end{figure*}

We train and validate our framework using a combination of several datasets.
Specifically, we merge different Copernicus datasets, as detailed next. First,  Sentinel-2 L2A cloud-free mosaics, using the whole 12-band input. Second, Corine Land Cover (CLC), which is a pan-European land cover classification dataset that provides information on land cover and land use. Third, the Land Use and Coverage Area frame Survey (LUCAS), a European-wide survey that collects data on land use and land cover across the continent. LUCAS provides only point-wise annotations, but the available areas have been manually validated, and therefore it is useful for both training and validation purposes. Fourth, Urban Atlas (UA), another European land cover and land use dataset localized around large urban areas, featuring higher spatial resolution, albeit with a reduced number of classes. Given the precise delineations offered by this source, we exploit it for training and validation purposes.
Fifth, the Dominant Leaf Type High-Resolution Layer (HRL), which provides information about the dominant leaf type across Europe.
Because our focus is on the production of fuel maps, we focus our study on the Mediterranean area, where wildfires are more frequent and intense. 
The ground truth consists of two annotations for each training region: a \textit{scribble} label, comprising a sparse fuel map derived from CLC, and a \textit{point-wise} label, derived from LUCAS. These annotations are the result of the following preprocessing pipelines.
Both the \textit{scribble} and the \textit{point-wise} labels are obtained by mapping the original classes into our fuel map taxonomy, as described in Table \ref{table:mapping}.

Following the methodology used to produce the S2GLC dataset \cite{s2glc2020malinowski}, we apply to the remapped CLC classes a filtering process using Normalized Difference Vegetation Index (NDVI) and Normalized Difference Water Index (NDWI) thresholds, eliminating potentially mislabeled pixels.
Next, we transform the filtered CLC classes into scribbles through morphological skeletonization followed by a small buffering of 5 pixels to increase their thickness.
Finally, urban category labels from CLC are replaced with more accurate UA labels, while we use the HRL dataset to differentiate wooded areas into coniferous and broadleaf forests, providing a more detailed categorization.
To generate the \textit{point-wise}, we rasterize the LUCAS points within the considered areas, assigning the LUCAS class to the closest fuel class.
These preprocessing steps generate a sparse ground truth with detailed fuel type information.


%% file: tables/mapping.tex
\definecolor{artificial}{RGB}{214,58,61}
\definecolor{bare}{RGB}{154,154,154}
\definecolor{wetlands}{RGB}{150,107,196}
\definecolor{water}{RGB}{43,80,198}
\definecolor{grassland}{RGB}{249,159,39}
\definecolor{agricultural}{RGB}{253,211,39}
\definecolor{broadleaves}{RGB}{36,152,1}
\definecolor{coniferous}{RGB}{8,98,0}
\definecolor{shrubs}{RGB}{141,140,0}
\definecolor{ignored}{RGB}{44,44,44}

\begin{adjustbox}{width=1.0\columnwidth}
\begin{tabular}{lccc} 
\midrule
\multicolumn{1}{c}
{\textbf{Fuel Class}} & \multicolumn{1}{c}
{\textbf{CLC Class ID}} & \multicolumn{1}{c}
{\textbf{LUCAS Class ID}} &
\multicolumn{1}{c}
{\textbf{Color}}\\ 
\midrule
Artificial & 111 112 121 122 123 124 131 132 133 142 & 7 & \cellcolor{artificial!50}\\
Bare & 331 332 335 & 6 & \cellcolor{bare!50}\\
Wetlands & 411 412 421 422 423 & & \cellcolor{wetlands!50}\\
Water & 511 512 521 522 523 & 8 9 & \cellcolor{water!50}\\
Grassland & 211 231 321 & 3 & \cellcolor{grassland!50}\\
Agricultural & 212 213 221 222 223 241 242 243 244 & 1 2 & \cellcolor{agricultural!50}\\
Broadleaves & 311 & 4 & \cellcolor{broadleaves!50}\\
Coniferous & 312 & 4 & \cellcolor{coniferous!50}\\
Shrubs & 322 323 324 333 & 5 & \cellcolor{shrubs!50}\\
Ignored & 141 313 334 999 & & \cellcolor{ignored!50}\\
\end{tabular}
\end{adjustbox}

%% file: sections/04-method.tex
\subsection{Problem statement}
We investigate a semantic segmentation task for fuel mapping in the presence of sparse annotations, a situation where only a subset of the pixels in an image are annotated with their corresponding class label, and the rest are left unmarked.
Let us define as $\mathcal{X}$ the set of multi-spectral input images, where each image $x$ is constituted by a set of pixels $\mathcal{I}$, and as $\mathcal{Y}$ the set of semantic annotations associating a class from the label set $\mathcal{C}$ to each pixel $j \in \mathcal{J}$, where $|\mathcal{J}| \ll |\mathcal{I}|$.
As described in Sec. \ref{sec:dataset}, we have two sets of sparsely-annotated maps: (i) a set of \textit{scribble} annotations, denoted as $Y_{S}$, and (ii) a set of \textit{point-wise} annotations, denoted as $Y_{P}$.
The goal is to find a parametric function $f_\theta$ that maps a multi-spectral image to a pixel-wise probability, i.e., $f_\theta: \mathcal{}{X} \rightarrow \mathcal{R}^{|\mathcal{I}|\times|\mathcal{C}|}$, and evaluate it on unseen images.
The parameters of the model $\theta$ are tuned to minimize a sum of two different categorical cross-entropy losses, namely 
 $L_{seg} = L_{\text{S}}(\hat{y}, y_{M}) + \lambda L_{\text{P}}(\hat{y}, y_{P})$, where $\hat{y}$ is the predicted label, $\lambda$ represents a weighting factor, while $y_M$ represents the ground truth derived from mixing scribbles with expanded pseudo-labels, as detailed in Sec. \ref{sec:framework}.
\subsection{Framework}
\label{sec:framework}
SPADA is based on DAFormer, a self-training UDA framework which consists of a transformer-based encoder with a multilevel context-aware decoder.
We adapt this UDA framework by substituting the concept of target domain with sparse labels. In this case, source and target data belong to the same image, where a small portion is provided with ground truth labels, and the remaining pixels remain unlabeled. We simplify the original framework by substituting RCS with simple class weights without performance loss, and by removing FD due to its inapplicability to land cover classes.
The SPADA framework is composed of three main blocks: (i) a \textit{student} network, trained on a mix of ground truth scribbles and dense pseudo-labels, and regularized using the \textit{point-wise} annotations, (ii) a \textit{teacher} network, obtained as EMA from the student model, that generates pseudo-labels from Sentinel-2 inputs in a robust and consistent way. Lastly, (iii) a \textit{label mixing} strategy between scribbles and these pseudo-labels.

The final training labels are obtained in two steps: first, the pseudo-labels are filtered based on a fixed confidence threshold, second the scribbles are fused on top of the remaining labels for consistency. Formally, each mixed label $y_M$ is obtained as composition of $\hat{y}_{T} \odot y_S$, where the $\hat{y}_{T}$ represents the pseudo-labels inferred by the teacher model, and $y_S$ identifies the \textit{scribble} annotations.
In order to avoid overconfident predictions, we include a weight on the mixed labels $w_i \in [0, 1]$ for each pixel $i$, where $w_i = 1$ if the pixel belongs to $y_S$, or $w_i = |\hat{\mathcal{I}}| / | \mathcal{I} |$ if $i$ belongs to $\hat{y}_{T}$. Here, $\hat{\mathcal{I}}$ is the set of pseudo-label pixels above a given threshold $\tau$.

\begin{figure}
    \centering
    \includegraphics[width=\linewidth]{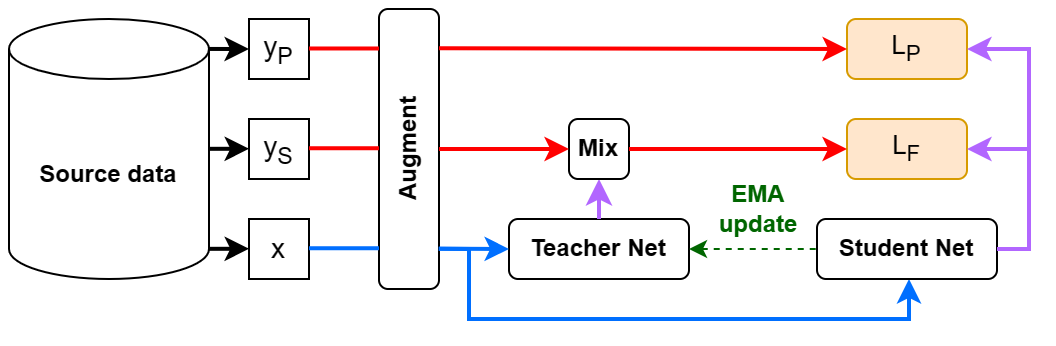}
    \caption{ Overview of SPADA framework.}
    \label{fig:spada}
\end{figure}

%% file: sections/05-experiments.tex
To cope with limited computational resources, we perform our evaluation over a total surface of $472,717Km^2$, encompassing the south European countries that have been most affected by wildfires in the last three years (i.e., Portugal, Spain, France, Italian regions, Greece, and Balkans). We use as input the Sentinel-2 cloudless mosaics, computed from April to September 2018. We select this interval to match the ground truth used (i.e., CLC, UA, VHR, LUCAS) and to enable a meaningful performance comparison with state-of-the-art products such as S2GLC.
We split the dataset into 12 equivalently sized areas, selecting 8 areas for training, and 4 for testing. Training areas are further split into training and validation sections of size $2,048 \times 2,048$ pixels, keeping 90\% and 10\% for training and validation respectively. All data is then tiled into $512 \times 512$ chips. The final set consists of $20,398$ tiles for training, 5100 for validation, and 394 full sections for testing that are divided into tiles at runtime.
We test our solution against semantic segmentation baselines and third-party products, namely the S2GLC \cite{s2glc2020malinowski} and the original Corine Land Cover. All the baseline models are trained on the fuel maps without sparse annotations, while CLC and S2GLC are only remapped to match the considered fuel classes.
Given the lack of dense annotations, we assess the performance of our solution on test areas using the two most reliable ground truths: LUCAS, using F1 score, and Urban Atlas, by means of the Intersection over Union (IoU) metric.
Each model is trained for $160,000$ iterations with an AdamW optimizer and a polynomial scheduler with a linear warm-up.
We further augment the inputs using horizontal and vertical flips, affine transforms, Gaussian blur, and we exploit test-time augmentations to improve the inference quality further.
We first evaluate the classification abilities of our system, comparing it to the manually validated LUCAS points. The results, listed in Table \ref{table:lucas}, show a consistent improvement of our model over all tested baselines, including Segformer, achieving $+3.86$ increment in terms of F1 score against S2GLC.
In Table \ref{table:ua}, we report the results in terms of IoU over the considered UA regions available in the test set. While specific categories (e.g., \textit{bare}, \textit{grassland}) are slightly underperforming, SPADA obtains on average comparable or substantially higher results on the other classes, with an IoU increment of $+7.88$ w.r.t. S2GLC.
Additionally, we include in both tables the performances computed on the raw CLC layers, which is the reference land cover product in Europe, showing that SPADA achieves better performances by a large margin.

\input{tables/LUCAS}
\input{tables/UA}
\begin{figure*}
\centering
\subfloat[Sentinel-2]{
\label{fig:qualitative_s2}
\centering
\includegraphics[width=0.22\textwidth]{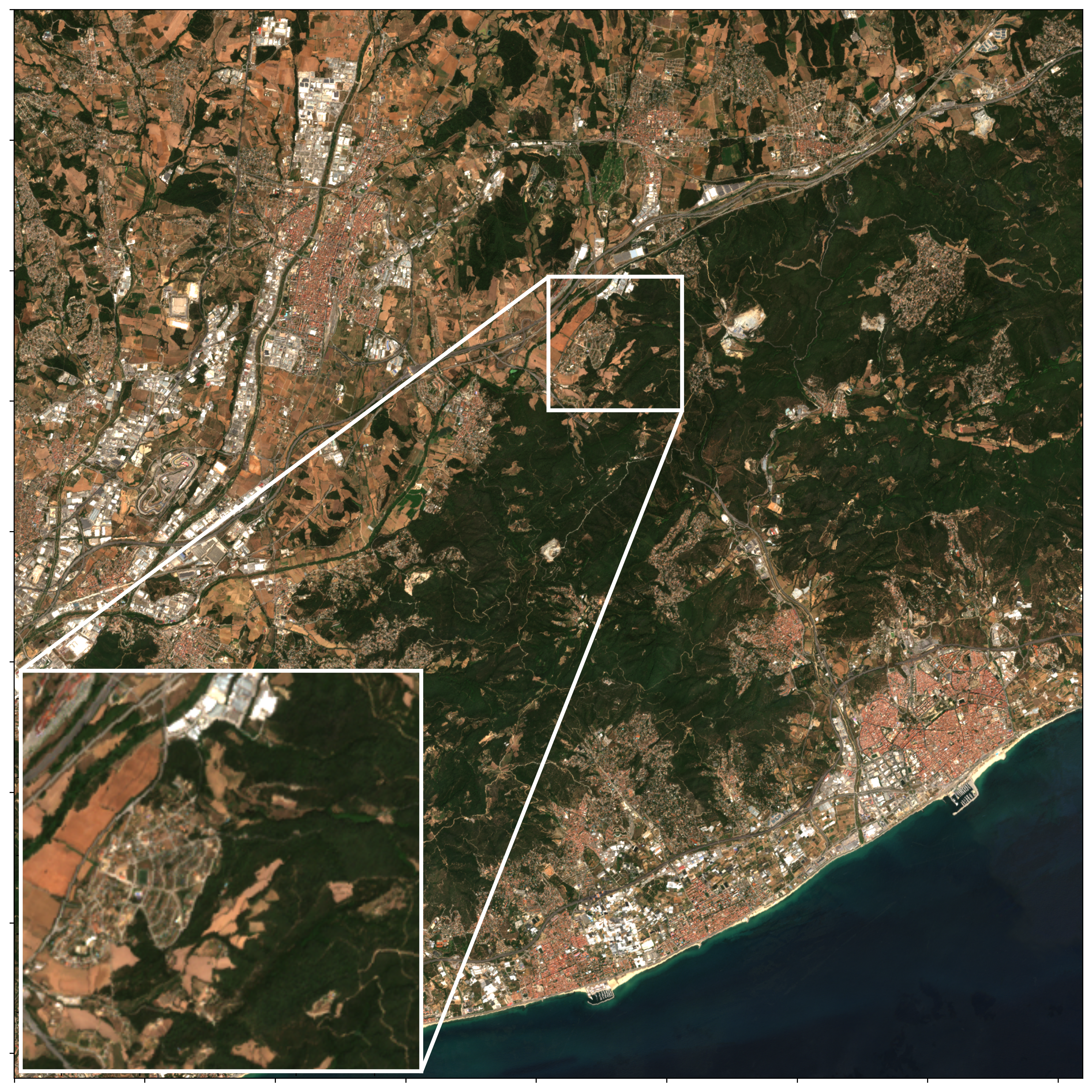}
}
\hfill
\subfloat[UNet baseline]{
\label{fig:qualitative_unet}
\centering
\includegraphics[width=0.22\textwidth]{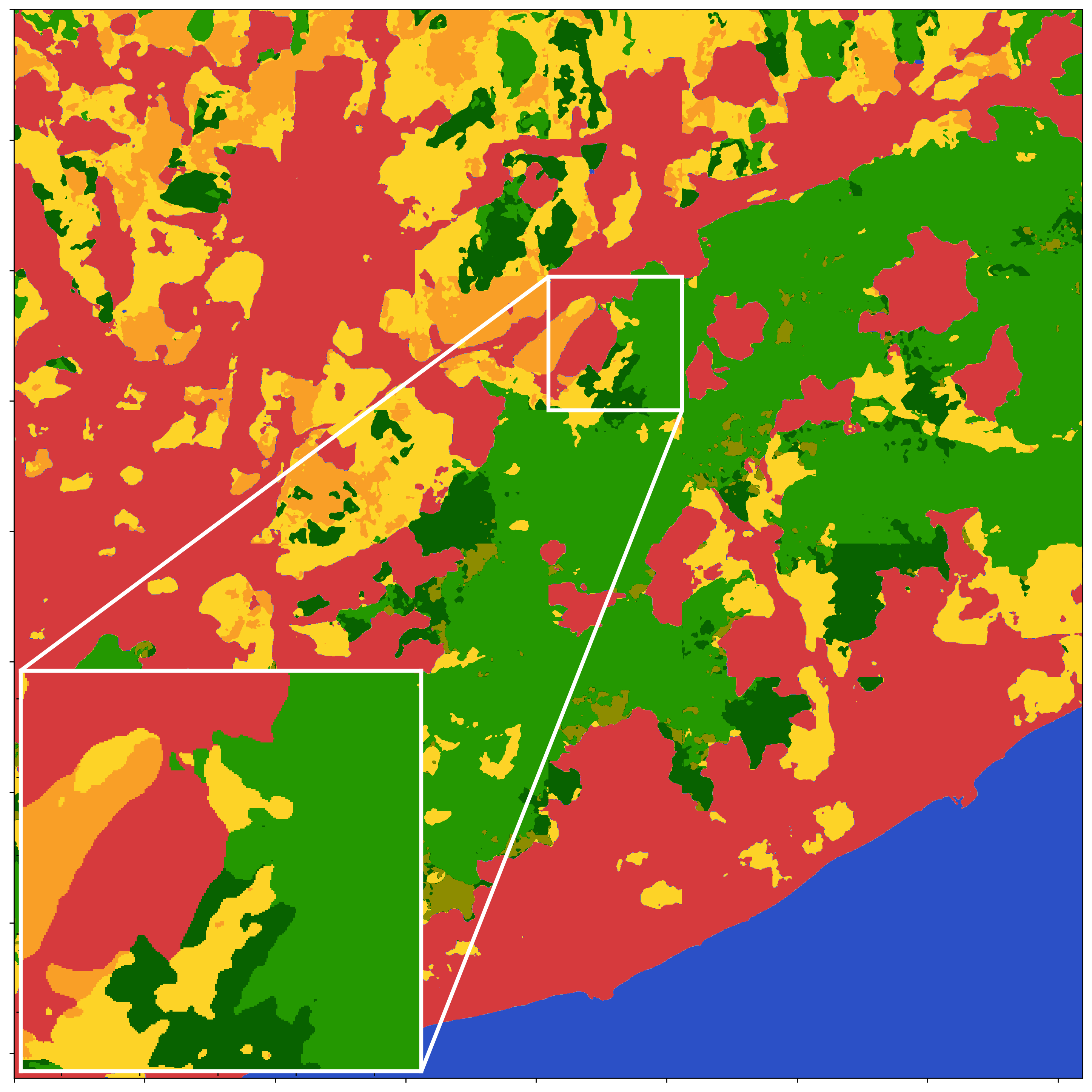}
}
\hfill
\subfloat[S2GLC]{
\label{fig:qualitative_s2glc}
\centering
\includegraphics[width=0.22\textwidth]{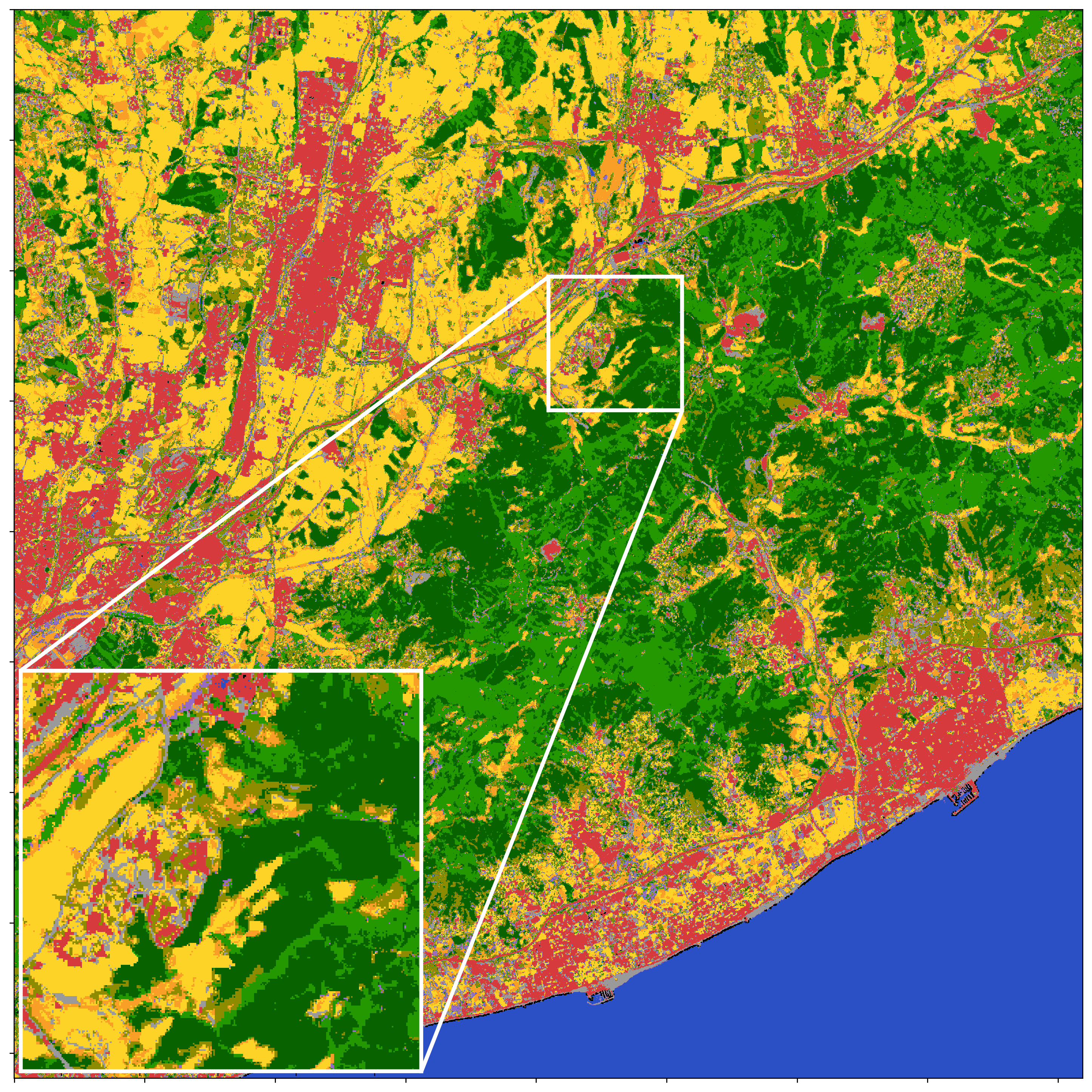}
}
\hfill
\subfloat[SPADA]{
\label{fig:qualitative_spada}
\centering
\includegraphics[width=0.22\textwidth]{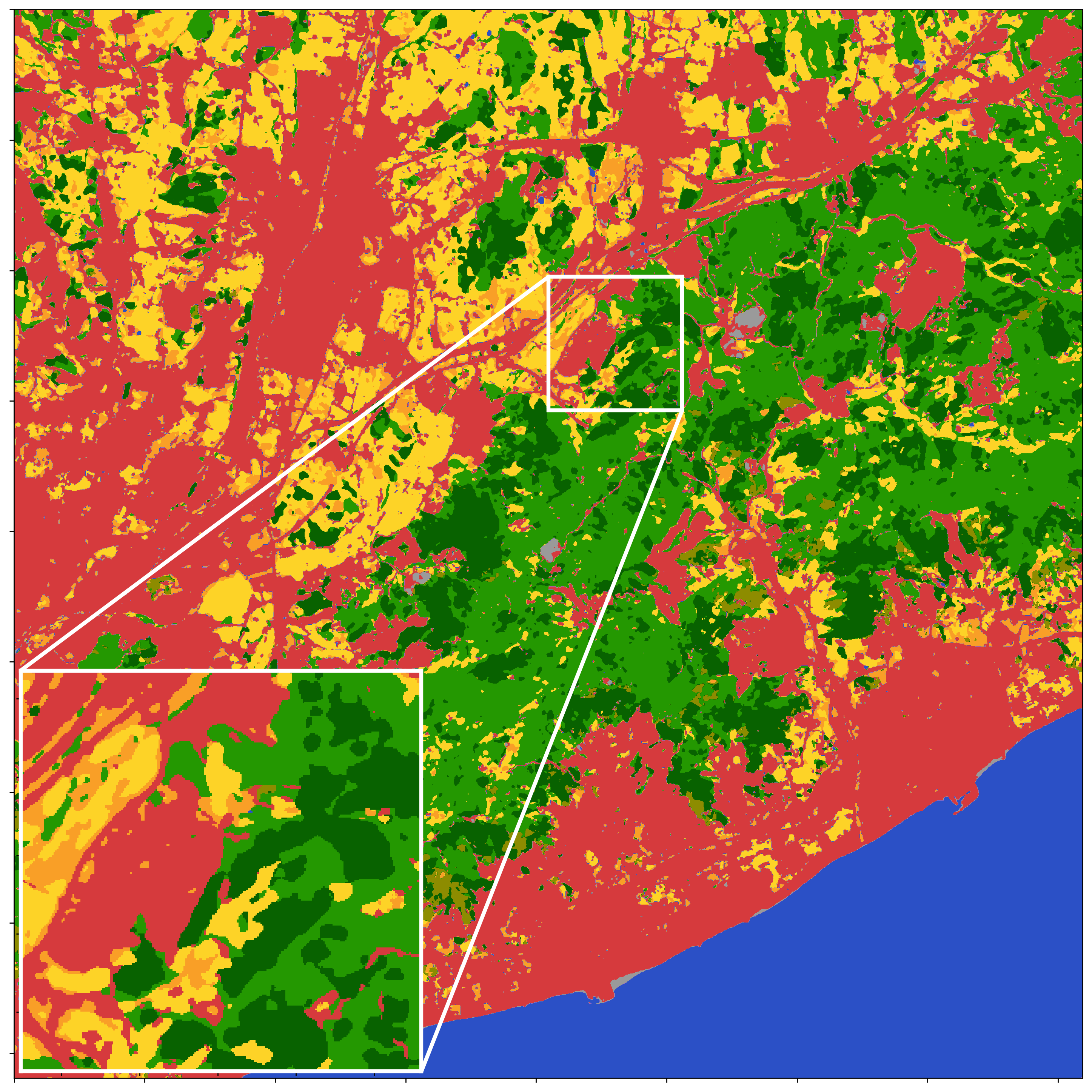}
}
\caption{Qualitative comparison between outputs, from left to right: S2 input, UNet baseline, S2GLC, SPADA. Best viewed zoomed in.}
\label{fig:qualitative}
\end{figure*}



%% file: tables/lucas.tex
\begin{table}[t]
\centering
\begin{adjustbox}{width=1.0\columnwidth}

\begin{tabular}{lrrrrrrrrrr} 
\midrule
\multicolumn{1}{c}{\textbf{Method}} & \multicolumn{1}{c}{\begin{sideways}\textbf{Agricultural}\end{sideways}} & \multicolumn{1}{c}{\begin{sideways}\textbf{Grassland}\end{sideways}} & \multicolumn{1}{c}{\begin{sideways}\textbf{Broadleaves}\end{sideways}} & \multicolumn{1}{c}{\begin{sideways}\textbf{Coniferous}\end{sideways}} & \multicolumn{1}{c}{\begin{sideways}\textbf{Shrubs}\end{sideways}} & \multicolumn{1}{c}{\begin{sideways}\textbf{Bare}\end{sideways}} & \multicolumn{1}{c}{\begin{sideways}\textbf{Artificial}\end{sideways}} & \multicolumn{1}{c}
{\begin{sideways}\textbf{Water}\end{sideways}} & \multicolumn{1}{c}
{\textbf{Avg}} & \multicolumn{1}{c}
{\textbf{Acc}}\\ 
\midrule
\rowcolor{lightgray} CLC & 55.72 & 24.71 & 66.49 & 68.51 & 33.55 & 60.63 & 52.71 & 63.40 & 55.53 & 52.54 \\
UNet & 57.94 & 28.40 & 72.30 & 68.33 & 34.72 & 34.47 & 55.98 & 73.01 & 58.00 & 55.62 \\
OCRNet & 59.39 & 26.25 & 70.71 & 65.17 & 33.77 & 33.81 & 53.88 & 68.73 & 56.87 & 54.38 \\
PSPNet & 58.51 & 18.69 & 66.38 & 64.56 & 31.66 & 52.87 & 52.88 & 68.18 & 55.04 & 52.10 \\
DeepLabV3Plus & 60.64 & 17.30 & 67.56 & 63.62 & 32.25 & 55.45 & 54.39 & 67.06 & 55.84 & 53.42 \\
SegFormer & 64.09 & 14.83 & 72.65 & 68.38 & 33.56 & 56.08 & 49.45 & 64.59 & 58.63 & 57.25 \\
S2GLC & 69.39 & 36.06 & 75.22 & 70.60 & 34.15 & 61.62 & 55.71 & \textbf{75.99} & 64.07 & 62.14 \\
\midrule
\textbf{SPADA (Ours)} & \textbf{77.72} & \textbf{39.36} & \textbf{76.78} & \textbf{74.19} & \textbf{38.08} & \textbf{63.64} & \textbf{58.54} & 70.49 & \textbf{67.93} & \textbf{66.99}
\end{tabular}

\end{adjustbox}
\caption{Experiments on LUCAS test set (F1 score).}
\label{table:lucas}
\end{table}

%% file: tables/ua.tex
\begin{table}[t]
\centering
\begin{adjustbox}{width=1.0\columnwidth}

\begin{tabular}{lrrrrrrrrr} 
\midrule
\multicolumn{1}{c}{\textbf{Method}} & \multicolumn{1}{c}{\begin{sideways}\textbf{Artificial}\end{sideways}} & \multicolumn{1}{c}{\begin{sideways}\textbf{Bare}\end{sideways}} & \multicolumn{1}{c}{\begin{sideways}\textbf{Wetlands}\end{sideways}} & \multicolumn{1}{c}{\begin{sideways}\textbf{Water}\end{sideways}} & \multicolumn{1}{c}{\begin{sideways}\textbf{Grassland}\end{sideways}} & \multicolumn{1}{c}{\begin{sideways}\textbf{Agricultural}\end{sideways}} & \multicolumn{1}{c}{\begin{sideways}\textbf{Forest}\end{sideways}} & \multicolumn{1}{c}
{\textbf{mIoU}} & \multicolumn{1}{c}
{\textbf{mAcc}}\\ 
\midrule
\rowcolor{lightgray} CLC & 58.86 & \textbf{14.78} & 32.44 & 58.68 & \textbf{14.62} & 36.56 & 45.69 & 37.37 & 49.81 \\
UNet & 51.06 & 1.83 & 25.10 & 32.00 & 11.77 & 39.88 & 53.67 & 30.76 & 49.04 \\
OCRNet & 53.15 & 2.39 & \textbf{34.36} & 30.79 & 12.57 & 39.34 & 50.59 & 31.88 & 48.15 \\
PSPNet & 55.12 & 3.85 & 34.12 & 30.68 & 11.81 & 38.72 & 46.57 & 31.55 & 47.64 \\
DeepLabV3Plus & 54.09 & 5.6 & 30.06 & 31.25 & 10.74 & 41.31 & 48.45 & 31.64 & 47.55 \\
SegFormer & 59.06 & 13.01 & 24.46 & 52.9 & 8.52 & 43.56 & 52.17 & 36.24 & 53.82 \\
S2GLC & 40.55 & 7.12 & 4.00 & \textbf{66.97} & 14.19 & 46.09 & \textbf{65.95} & 34.98 & 52.49 \\
\midrule
\textbf{SPADA (Ours)} & \textbf{64.36} & 13.56 & 27.27 & 65.95 & 10.01 & \textbf{54.3} & 64.56 & \textbf{42.86} & \textbf{58.11}
\end{tabular}

\end{adjustbox}
\caption{Experiments on Urban Atlas test set (IoU).}
\label{table:ua}
\end{table}

%% file: sections/06-conclusions.tex
Exploiting a curated set of sparse annotations, we build an ad-hoc dataset for fuel map segmentation.
We then propose SPADA, a framework for sparsely annotated semantic segmentation inspired by UDA techniques. We perform an extensive performance evaluation of our framework over a wide area in Europe, showing that our solution outperforms both semantic segmentation baselines and existing land cover products such as CLC and S2GLC.
Future works will focus on expanding the available data with a wider range of geographical areas and modalities, as well as improving the methodology with ad-hoc refinements over the pseudo-label generation.